\documentclass[11pt]{article}

\usepackage[final]{acl}

\usepackage{times}
\usepackage{latexsym}

\usepackage[T1]{fontenc}

\usepackage[utf8]{inputenc}

\usepackage{microtype}
\usepackage {placeins}

\usepackage{inconsolata}

\usepackage{graphicx}

\usepackage{amsmath}
\usepackage{amssymb}
\usepackage{booktabs}
\usepackage{algorithm}
\usepackage{algpseudocode}
\usepackage{multirow}
\usepackage{xspace}
\usepackage{enumitem}
\usepackage{caption}
\usepackage{tikz}
\usetikzlibrary{arrows.meta, positioning, fit, backgrounds, calc, shapes.geometric, decorations.pathreplacing}
\usepackage{pgfplots}
\pgfplotsset{compat=1.18}

\newcommand{\methodname}{RECAP\xspace}
\newcommand{\benchmarkexpansion}{Regression Evaluation for Continual Adaptation of Prompts}

\newcommand{\satrate}{r_{\text{sat}}}

\newcommand{\peakforg}{f_{\text{peak}}}

\newcommand{\uf}{\text{UF}}

\title{\methodname: \benchmarkexpansion}

\author{
  \textbf{Harsh Deshpande} \quad
  \textbf{Kushal Chawla} \quad
  \textbf{Sangwoo Cho} \\
  \textbf{William Campbell} \quad
  \textbf{Sambit Sahu} \\
  Capital One \\
  \texttt{\{harsh.deshpande2,kushal.chawla\}@capitalone.com}
}

\begin{document}
\maketitle
\begin{abstract}
Production agentic systems routinely face evolving constraints and must comply from the very next interaction. Scenarios like a tool-call notification changing a compliance threshold or a policy update adding disclosure requirements fit this criteria, having close to no room for errors in production.
This \emph{proactive} adaptation setting is common in deployment, but absent from current benchmarks, which assume either static constraint sets or reactive protocols with evaluation feedback.
We introduce \methodname, a benchmark that measures continual-learning phenomena (forgetting, regression, forward transfer) at the constraint level under a strictly proactive adapt-then-test protocol: prompt optimization methods receive only the constraint specification and must generalize before seeing any test data. Evaluating six methods across five LLMs and three schedules with evolving constraints, we find that these methods show no significant improvement in performance, even after incurring a higher latency. These methods, designed for offline or reactive settings, are inadequate for the proactive paradigm. Our work emphasizes the growing need for designing proactive prompt adaptation methods, where the models must remain robust to evolving needs in deployment.
\end{abstract}

\section{Introduction}
\label{sec:intro}

Agentic systems in production operate under constraints that evolve continuously: a tool-call response tightens a length limit, a policy update adds a disclosure requirement, or a user preference changes the expected tone.
The system must satisfy new constraints immediately while continuing to respect all prior ones. Further, these constraints are often not centrally documented: they accumulate from individual tool-call responses, user interactions, and personalization settings across many users \citep{CCTU}, making it unrealistic to collect the full active set and optimize jointly each time one changes \citep{banerjee2025crane}.

Aligned with the EMNLP 2026 theme of rethinking evaluation beyond static benchmarks, we argue that progress in deployed agents must account for longitudinal behavior under evolving specifications.
We focus on \textit{proactive} adaptation, where the system receives only a constraint specification and must comply before seeing any real test data or feedback, with minimal latency overhead. The setting is ubiquitous in deployment, yet entirely absent from current evaluation paradigms.
Instruction-following benchmarks \citep{zhou2023ifeval, guo2025recast, jiang2024followbench, qin2024infobench} present a fixed constraint set and measure single-shot success, with no mechanism for constraints to evolve over time.
Prompt optimization methods \citep{yuksekgonul2025textgrad, yang2023opro, khattab2024dspy, opsahlong2024miprov2} do address iterative improvement, but assume access to representative evaluation data and multiple rounds of feedback, which are not available in the proactive setting.
Reactive protocols like ACE \citep{zhang2025agenticcontextengineeringevolving} allow iterative debugging on observed failures, but again require test-time feedback to drive adaptation.
The natural framework for evolving constraints is Continual learning (CL), which studies how systems adapt to new tasks without forgetting prior ones \citep{delange2021clsurvey, ke2024continual, wu2024continual}.
However, existing CL operate on model weights through regularization \citep{kirkpatrick2017ewc}, replay \citep{lopezpaz2017gem}, or prompt embeddings \citep{wang2022learning, wang2022dualprompt} and do not address prompt-level text constraints where the model weights are frozen and adaptation must happen entirely through the input. None evaluates the proactive case where methods must generalize from specification alone.

We present \textbf{RECAP}: \textbf{R}egression \textbf{E}valuation for \textbf{C}ontinual \textbf{A}daptation of \textbf{P}rompts, a benchmark that extends the evaluation of constraint satisfaction beyond static settings. RECAP performs continual evaluation on schedules subject to evolving constraints with \textit{add}, \textit{edit}, and \textit{delete} operations. This enables a rigorous evaluation of recent prompt adaptation methods under a proactive protocol. We summarize our contributions below:


\begin{enumerate}[nosep,leftmargin=*]
\item We design a \textbf{constraint-level CL benchmark} that converts static instruction-following datasets into temporal evaluation streams via typed operations under a proactive protocol, where methods receive only the constraint specification and must generalize immediately (\S\ref{sec:method}).
\item We develop a \textbf{decomposed metric suite} measuring constraint satisfaction, regression, edit uptake (\textit{Are modified constraints adopted?}), unlearning fidelity (\textit{Are deleted constraints forgotten?}), and efficiency (\S\ref{sec:method}).
\item We provide \textbf{empirical evidence} that existing prompt adaptation methods are structurally inadequate with the proactive paradigm, discussing failure modes to guide future work (\S\ref{sec:experiment} and \S\ref{sec:results}).
\end{enumerate}

\section{Methodology}
\label{sec:method}
%
\noindent\textbf{Source Data:} In the proactive setting, constraints evolve independently of the base task. This requires source data where constraints are separated from base instructions.
We build on RECAST-30K \citep{guo2025recast}: which is based on Tulu~3 Persona IF \citep{lambert2024tulu3}. The data contains base instructions (e.g., `\textit{Write a cover letter for a data-science role}') paired with one or more constraints (e.g., `\textit{keep under 200 words}' or `\textit{mention Python at least 3 times}'). Constraints are grouped into semantic \emph{types} (Length, Keyword, Format, Tone, etc.), each with one or more concrete values (e.g. maximum length can be 200 or 300 words). We have 21 constraints in total, out of which 8 have deterministic rule-based validators while 13 require LLM judgement (Appendix~\ref{app:taxonomy}).
%

\noindent\textbf{Operations and Shadow Evaluation:} \methodname transforms this static dataset of instructions and constraints into a temporal evaluation stream by defining \textit{schedules} of evolving constraints. At each step in a schedule, we apply one out of three operations: 1) \textsc{Add} introduces a new constraint type, 2) \textsc{Edit} replaces the concrete value of an existing type, and 3) \textsc{Delete} removes a constraint type entirely. A \emph{schedule} consists of a sequence of operations over 15-20 steps (Appendix~\ref{app:config}), controlling which constraints are introduced, modified, or removed, and in what order. A key question is whether adapting to a new or modified constraint causes interference with previously satisfied ones. To measure this after edits and deletions, we retain the old constraint as a \textit{shadow} in the evaluation set at all subsequent steps: responses continue to be checked against the replaced or removed specification even though the LLM no longer sees it.
This enables tracking of edit persistence (\textit{does the model revert to old behavior over time?}) and unlearning rebound (\textit{does a deleted constraint resurface?}).

\noindent\textbf{RECAP Protocol:} We adopt the adapt-then-test protocol from CL evaluation \citep{lopezpaz2017gem, chaudhry2018riemannian, chaudhry2019agem, delange2021clsurvey}.
At each step, the method first adapts to the new constraint operation, then is evaluated on all active constraints (see Figure~\ref{fig:protocol}, pseudocode is in Appendix~\ref{app:protocol}).
Adaptation is \emph{proactive}: \texttt{adapt()} receives only the constraint specification (e.g., ``edit Length: Keep under 500 words'') but no test prompts and no feedback from evaluation.
Methods may use internal self-play during adaptation (generating and judging synthetic responses), but they never observe real test data or evaluation results from prior steps. The no-adaptation baseline (Base LLM) skips adaptation entirely and receives only the current active constraints appended to each user prompt at test time, making it a pure test of the LLM's instruction-following ability.

\begin{figure*}[t]
\centering
\resizebox{0.97\textwidth}{!}{%
\begin{tikzpicture}[
    >=Stealth,
    card/.style={fill=white, rounded corners=6pt,
        minimum width=3.1cm,
        text width=2.65cm, align=left, font=\scriptsize, inner sep=6pt},
    prodstep/.style={card, draw=blue!28, minimum height=2.45cm},
    recapstep/.style={card, draw=green!35!black, minimum height=3.75cm},
    paneltitle/.style={font=\scriptsize\bfseries, text width=1.55cm,
        align=right, inner sep=0pt},
    maparrow/.style={->, densely dotted, gray!55, line width=0.65pt},
    timearrow/.style={-{Stealth[length=1.7mm]}, gray!50, line width=0.7pt},
]

\newcommand{\activeconstraintsbox}[1]{%
  \begingroup
  \setlength{\fboxsep}{2pt}%
  \setlength{\fboxrule}{0.35pt}%
  \fcolorbox{green!30}{green!3}{%
    \parbox[t][0.82cm][t]{2.28cm}{\vspace{0pt}\raggedright
      \textcolor{green!45!black}{\fontsize{5.2}{5.8}\selectfont
      \textbf{Active constraints}}\\[-1pt]
      {\fontsize{6}{6.8}\selectfont #1}}}%
  \endgroup
}
\newcommand{\shadowconstraintsbox}[1]{%
  \begingroup
  \setlength{\fboxsep}{2pt}%
  \setlength{\fboxrule}{0.35pt}%
  \fcolorbox{gray!35}{gray!4}{%
    \parbox[t][0.82cm][t]{2.28cm}{\vspace{0pt}\raggedright
      \textcolor{black!78}{\fontsize{5.2}{5.8}\selectfont
      \textbf{Shadow evaluation constraints}}\\[-1pt]
      {\fontsize{6}{6.8}\selectfont #1}}}%
  \endgroup
}
\newcommand{\productiondetail}[2]{%
  \parbox[t][1.05cm][t]{2.65cm}{\vspace{0pt}\raggedright
    \textbf{#1}\\[1pt]
    \textcolor{blue!70!black}{#2}}%
}
\newcommand{\recapoperation}[1]{%
  \parbox[t][0.36cm][t]{2.65cm}{\vspace{0pt}\texttt{#1}}%
}

\node[prodstep] (p1) {
    {\color{blue!35!black}\bfseries UPDATE 1}\hfill
    {\color{blue!65!black}\bfseries TOOL}\\[3pt]
    \productiondetail{Tool notification}{New limit: $<300$ words}\\[4pt]
    \colorbox{orange!12}{\strut adapt}
    $\;\to\;$
    \colorbox{green!12}{\strut serve}
};

\node[prodstep, right=0.32cm of p1] (p2) {
    {\color{blue!35!black}\bfseries UPDATE 2}\hfill
    {\color{blue!65!black}\bfseries USER}\\[3pt]
    \productiondetail{User preference}{Use a formal tone}\\[4pt]
    \colorbox{orange!12}{\strut adapt}
    $\;\to\;$
    \colorbox{green!12}{\strut serve}
};

\node[prodstep, right=0.32cm of p2] (p3) {
    {\color{blue!35!black}\bfseries UPDATE 3}\hfill
    {\color{blue!65!black}\bfseries POLICY}\\[3pt]
    \productiondetail{Policy revision}{Limit changes to $<500$}\\[4pt]
    \colorbox{orange!12}{\strut adapt}
    $\;\to\;$
    \colorbox{green!12}{\strut serve}
};

\node[prodstep, right=0.32cm of p3] (p4) {
    {\color{blue!35!black}\bfseries UPDATE 4}\hfill
    {\color{blue!65!black}\bfseries USER}\\[3pt]
    \productiondetail{Preference removed}{Formal tone no longer required}\\[4pt]
    \colorbox{orange!12}{\strut adapt}
    $\;\to\;$
    \colorbox{green!12}{\strut serve}
};

\node[paneltitle, text=blue!65!black, left=0.38cm of p1]
    {(a) Production\\deployment};
\draw[timearrow] (p1) -- (p2);
\draw[timearrow] (p2) -- (p3);
\draw[timearrow] (p3) -- (p4);

\node[recapstep, below=1.55cm of p1] (r1) {
    {\color{blue!35!black}\bfseries STEP 1}\hfill
    {\color{blue!65!black}\bfseries ADD}\\[3pt]
    \recapoperation{+ Length $<300$}\\[5pt]
    \activeconstraintsbox{Length $<300$}\\[3pt]
    \shadowconstraintsbox{None}
};

\node[recapstep, right=0.32cm of r1] (r2) {
    {\color{blue!35!black}\bfseries STEP 2}\hfill
    {\color{blue!65!black}\bfseries ADD}\\[3pt]
    \recapoperation{+ Tone = formal}\\[5pt]
    \activeconstraintsbox{Length $<300$; Tone = formal}\\[3pt]
    \shadowconstraintsbox{None}
};

\node[recapstep, right=0.32cm of r2] (r3) {
    {\color{blue!35!black}\bfseries STEP 3}\hfill
    {\color{orange!80!black}\bfseries EDIT}\\[3pt]
    \recapoperation{Length $<300\to{}<500$}\\[5pt]
    \activeconstraintsbox{Length $<500$; Tone = formal}\\[3pt]
    \shadowconstraintsbox{Length $<300$}
};

\node[recapstep, right=0.32cm of r3] (r4) {
    {\color{blue!35!black}\bfseries STEP 4}\hfill
    {\color{red!65!black}\bfseries DELETE}\\[3pt]
    \recapoperation{- Tone = formal}\\[5pt]
    \activeconstraintsbox{Length $<500$}\\[3pt]
    \shadowconstraintsbox{Length $<300$; Tone = formal}
};

\node[paneltitle, text=green!40!black, left=0.38cm of r1]
    {(b) \methodname\\protocol};
\draw[timearrow] (r1) -- (r2);
\draw[timearrow] (r2) -- (r3);
\draw[timearrow] (r3) -- (r4);

\draw[maparrow] (p1.south) -- (r1.north);
\draw[maparrow] (p2.south) -- (r2.north);
\draw[maparrow] (p3.south) -- (r3.north);
\draw[maparrow] (p4.south) -- (r4.north);
\node[font=\scriptsize, text=blue!35!black, fill=white,
      rounded corners=2pt, inner sep=2pt]
      at ($(p2.south)!0.5!(r3.north)$)
      {\textit{each update becomes one proactive} \texttt{adapt()}$\to$\texttt{test()} \textit{step}};

\begin{scope}[on background layer]
\node[draw=blue!12, fill=blue!2, rounded corners=9pt,
      fit=(p1)(p2)(p3)(p4), inner xsep=8pt, inner ysep=8pt] {};
\node[draw=green!20, fill=green!2, rounded corners=9pt,
      fit=(r1)(r2)(r3)(r4), inner xsep=8pt, inner ysep=8pt] {};
\end{scope}

\node[draw=red!18, fill=red!3, rounded corners=4pt,
      below=0.3cm of r2.south east, font=\scriptsize, align=center,
      inner sep=4pt, text=red!65!black] (nofeedback)
      {\textbf{No test feedback:} test prompts and outcomes are never returned to \texttt{adapt()}.};

\end{tikzpicture}
}%
\caption{The \methodname protocol mirrors production deployment: a constraint specification arrives, the method adapts without access to test data, then is evaluated on all active constraints. This repeats over a multi-step schedule.}
\label{fig:protocol}
\end{figure*}
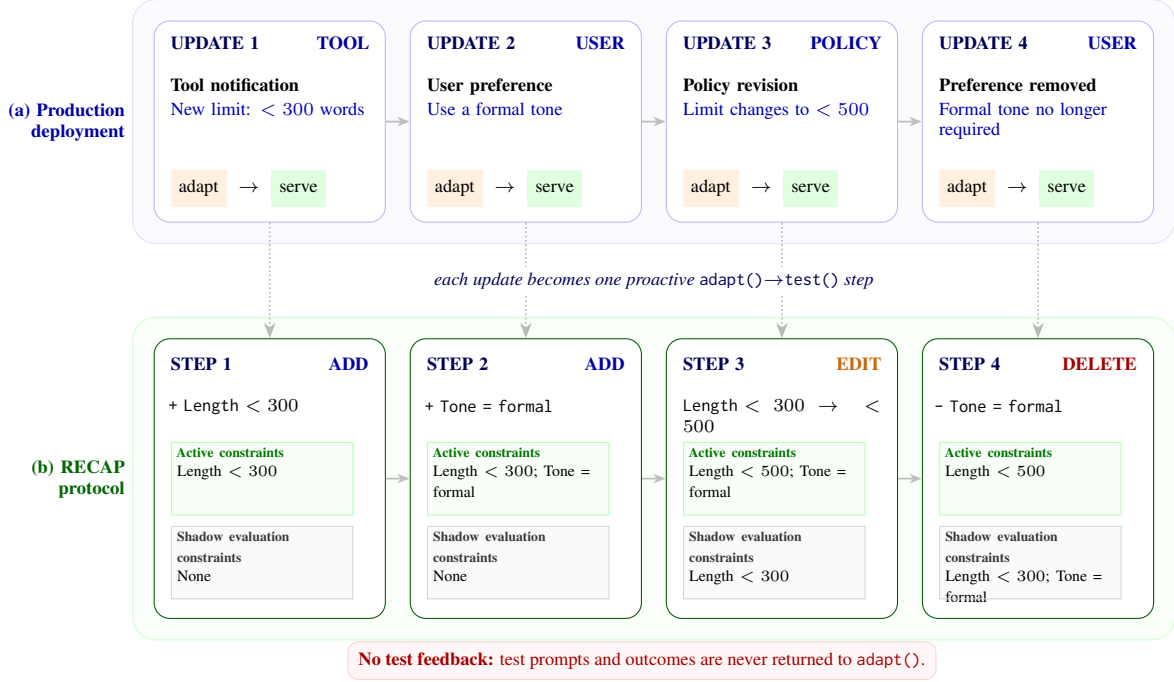

\noindent\textbf{Metrics:} Our primary metric is $\overline{\text{sat}}$: Mean constraint satisfaction rate across all types and steps (formula in Appendix~\ref{app:metrics}).
However, $\overline{\text{sat}}$ by itself can mask important dynamics. A method might maintain mean satisfaction while silently regressing on prior constraints -- We detect this with \emph{peak forgetting} \citep{chaudhry2018riemannian} (the maximum drop from a type's prior peak) and \emph{collateral damage} (mean drop in non-targeted types after an operation).
For edits, we define \emph{edit switch}: the fraction of samples satisfying the new specification but not the old.
For deletions, we ask whether the model appropriately stops satisfying a removed constraint: \emph{Unlearning Fidelity} measures how quickly satisfaction reverts to its unconstrained default rate. We also report latency (Appendices~\ref{app:perbackbone} and~\ref{app:gemma}).

\noindent\textbf{Methods:} We evaluate $6$ prompt adaptation methods, spanning few-shot (ICL), memory-based (Dynamic Cheatsheet \citep{suzgun2025dynamic}), and optimization-based (ACE, GEPA, MIPROv2 \citep{opsahlong2024miprov2}) paradigms. The optimization methods use \emph{self-play} during \texttt{adapt()}: the LLM generates a response to a synthetic prompt, then a second call judges whether the target constraint is satisfied. They differ in search strategy: ACE uses a generate--evaluate--reflect--curate pipeline (4 LLM calls/step); GEPA evolves a population via mutation and fitness selection (17 calls/step); MIPROv2 proposes diverse candidates informed by a score history (${\sim}$11 calls/step). To mimic dynamic realistic settings, all methods are asked to optimize for the single new constraint at each step without access to other active constraints.


\begin{figure*}[!t]
\centering
\begin{tikzpicture}
\begin{axis}[
    width=0.95\textwidth,
    height=6.7cm,
    ybar=2.2pt,
    bar width=7pt,
    ylabel={Mean Satisfaction},
    ylabel style={font=\small},
    symbolic x coords={Llama-8B, GPT-20B, Llama-70B, GPT-120B, Gemma-26B},
    xtick=data,
    xticklabel style={font=\small},
    yticklabel style={font=\small},
    ymin=0.25, ymax=0.72,
    ytick={0.3, 0.4, 0.5, 0.6, 0.7},
    legend style={
        font=\scriptsize,
        at={(0.5,0.97)},
        anchor=north,
        legend columns=3,
        column sep=4pt,
        /tikz/every even column/.append style={column sep=6pt},
    },
    error bars/y dir=both,
    error bars/y explicit,
    error bars/error bar style={line width=0.4pt},
    every axis plot/.append style={fill opacity=0.85},
    grid=major,
    grid style={gray!20, line width=0.3pt},
    major tick length=2pt,
    enlarge x limits=0.10,
]
\addplot[fill=gray!50, draw=gray!70] coordinates {
    (Llama-8B, 0.490) +- (0,0.020)
    (GPT-20B, 0.555) +- (0,0.016)
    (Llama-70B, 0.595) +- (0,0.011)
    (GPT-120B, 0.630) +- (0,0.023)
    (Gemma-26B, 0.677) +- (0,0.006)
};
\addplot[fill=blue!45, draw=blue!65] coordinates {
    (Llama-8B, 0.399) +- (0,0.052)
    (GPT-20B, 0.539) +- (0,0.022)
    (Llama-70B, 0.592) +- (0,0.016)
    (GPT-120B, 0.598) +- (0,0.023)
    (Gemma-26B, 0.627) +- (0,0.037)
};
\addplot[fill=green!45, draw=green!65] coordinates {
    (Llama-8B, 0.480) +- (0,0.042)
    (GPT-20B, 0.549) +- (0,0.036)
    (Llama-70B, 0.600) +- (0,0.010)
    (GPT-120B, 0.610) +- (0,0.036)
    (Gemma-26B, 0.678) +- (0,0.017)
};
\addplot[fill=orange!55, draw=orange!75] coordinates {
    (Llama-8B, 0.425) +- (0,0.106)
    (GPT-20B, 0.379) +- (0,0.106)
    (Llama-70B, 0.602) +- (0,0.014)
    (GPT-120B, 0.454) +- (0,0.151)
    (Gemma-26B, 0.504) +- (0,0.170)
};
\addplot[fill=red!40, draw=red!65] coordinates {
    (Llama-8B, 0.483) +- (0,0.034)
    (GPT-20B, 0.466) +- (0,0.053)
    (Llama-70B, 0.603) +- (0,0.004)
    (GPT-120B, 0.506) +- (0,0.100)
    (Gemma-26B, 0.658) +- (0,0.032)
};
\addplot[fill=purple!40, draw=purple!65] coordinates {
    (Llama-8B, 0.507) +- (0,0.021)
    (GPT-20B, 0.415) +- (0,0.068)
    (Llama-70B, 0.595) +- (0,0.009)
    (GPT-120B, 0.571) +- (0,0.113)
    (Gemma-26B, 0.660) +- (0,0.035)
};
\legend{Base LLM, ICL, Dyn.\ Cheatsheet, ACE, GEPA, MIPROv2}
\end{axis}
\end{tikzpicture}
\caption{Mean satisfaction ($\pm1$ SD across three schedules) for all six methods and five backbone LLMs.}
\label{fig:main_results}
\end{figure*}
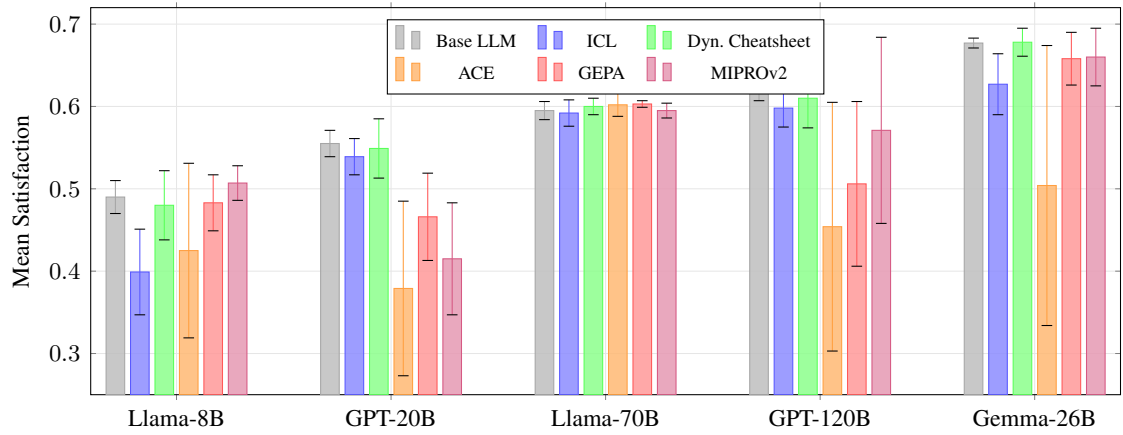

\section{Experimental Setup}
\label{sec:experiment}
We show aggregated results on $50$ base prompts taken from RECAST using 3 schedules, each designed to assess a different aspect of continual adaptation. \textbf{Interleaved-20} mixes 11 adds, 5 edits, and 4 deletes across 20 steps, testing whether methods handle concurrent accumulation and revision. \textbf{Clustered-20} applies the same operations but phased (ADD$\to$EDIT$\to$DELETE blocks), testing whether batched operations amplify forgetting. \textbf{Rule-Only-15} uses 6 rule-based types with deterministic validators in an interleaved structure, isolating genuine forgetting from any LLM judge noise (more details in Appendix~\ref{app:config}). We use $5$ backbone LLMs: Llama-3.1-8B, Llama-3.3-70B, GPT-OSS-20B, GPT-OSS-120B, and Gemma-4-26B-A4B \citep{grattafiori2024llama3,agarwal2025gpt,gemmateam2026gemma4technicalreport}. Claude Sonnet~4.5 \citep{anthropic2025claude} serves as the LLM judge for qualitative constraints. In total, we have 90 conditions (5 backbones $\times$ 6 methods $\times$ 3 schedules). Detailed hyperparameters and the judge protocol are in Appendix~\ref{app:judge_protocol}.

\section{Results}
\label{sec:results}

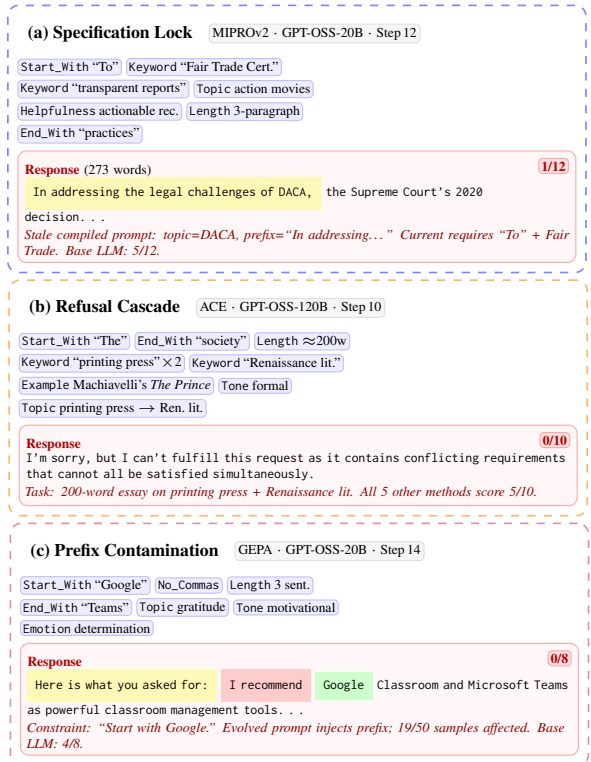
\begin{figure}[!htbp]
\centering
\resizebox{\columnwidth}{!}{%
\begin{tikzpicture}[
    panel/.style={draw=#1, dashed, rounded corners=4pt, line width=0.6pt, inner sep=3.5pt},
    ctag/.style={fill=blue!7, draw=blue!25, rounded corners=1.5pt, font=\fontsize{5}{6}\selectfont, inner sep=1pt},
    callout/.style={draw=red!35, fill=red!3, rounded corners=3pt, font=\fontsize{5}{6}\selectfont, text width=7cm, inner sep=2.5pt, align=left},
    badge/.style={fill=red!12, draw=red!35, rounded corners=1.5pt, font=\fontsize{5}{6}\selectfont\bfseries, inner sep=1pt, text=red!70!black},
    mlabel/.style={font=\fontsize{5}{6}\selectfont, fill=gray!7, draw=gray!25, rounded corners=1.5pt, inner sep=1.2pt},
    ptitle/.style={font=\fontsize{6.5}{7.5}\selectfont\bfseries},
]

\node[ptitle] (t1) at (-0.3, 0) {(a) Specification Lock};
\node[mlabel, right=2pt of t1] {MIPROv2 $\cdot$ GPT-OSS-20B $\cdot$ Step\,12};

\node[ctag, below=2.5pt of t1.south west, anchor=north west] (c1a) {\texttt{Start\_With} ``To''};
\node[ctag, right=1.5pt of c1a] (c1b) {\texttt{Keyword} ``Fair Trade Cert.''};
\node[ctag, below=1.5pt of c1a.south west, anchor=north west] (c1c) {\texttt{Keyword} ``transparent reports''};
\node[ctag, right=1.5pt of c1c] (c1d) {\texttt{Topic} action movies};
\node[ctag, below=1.5pt of c1c.south west, anchor=north west] (c1e) {\texttt{Helpfulness} actionable rec.};
\node[ctag, right=1.5pt of c1e] (c1f) {\texttt{Length} 3-paragraph};
\node[ctag, below=1.5pt of c1e.south west, anchor=north west] (c1g) {\texttt{End\_With} ``practices''};

\node[callout, below=2.5pt of c1g.south west, anchor=north west] (r1) {
\textbf{\color{red!70!black}Response} (273 words) \hfill \tikz\node[badge]{1/12};\\[-1pt]
{\ttfamily \colorbox{yellow!35}{\strut In addressing the legal challenges of DACA,} the Supreme Court's 2020 decision\ldots}\\[0.5pt]
{\color{red!55!black}\itshape Stale compiled prompt: topic=DACA, prefix=``In addressing\ldots'' Current requires ``To'' + Fair Trade. Base LLM: 5/12.}
};

\begin{scope}[on background layer]
\node[panel=blue!50, fit=(t1)(c1b)(c1g)(r1)] (p1) {};
\end{scope}

\node[ptitle, below=6pt of p1.south west, anchor=north west, xshift=0.15cm] (t2) {(b) Refusal Cascade};
\node[mlabel, right=2pt of t2] {ACE $\cdot$ GPT-OSS-120B $\cdot$ Step\,10};

\node[ctag, below=2.5pt of t2.south west, anchor=north west] (c2a) {\texttt{Start\_With} ``The''};
\node[ctag, right=1.5pt of c2a] (c2b) {\texttt{End\_With} ``society''};
\node[ctag, right=1.5pt of c2b] (c2c) {\texttt{Length} $\approx$200w};
\node[ctag, below=1.5pt of c2a.south west, anchor=north west] (c2d) {\texttt{Keyword} ``printing press''$\times$2};
\node[ctag, right=1.5pt of c2d] (c2e) {\texttt{Keyword} ``Renaissance lit.''};
\node[ctag, below=1.5pt of c2d.south west, anchor=north west] (c2f) {\texttt{Example} Machiavelli's \textit{The Prince}};
\node[ctag, right=1.5pt of c2f] (c2g) {\texttt{Tone} formal};
\node[ctag, below=1.5pt of c2f.south west, anchor=north west] (c2h) {\texttt{Topic} printing press $\to$ Ren.\ lit.};

\node[callout, below=2.5pt of c2h.south west, anchor=north west] (r2) {
\textbf{\color{red!70!black}Response} \hfill \tikz\node[badge]{0/10};\\[-1pt]
{\ttfamily I'm sorry, but I can't fulfill this request as it contains conflicting requirements that cannot all be satisfied simultaneously.}\\[0.5pt]
{\color{red!55!black}\itshape Task: 200-word essay on printing press + Renaissance lit. All 5 other methods score 5/10.}
};

\begin{scope}[on background layer]
\node[panel=orange!60, fit=(t2)(c2c)(c2h)(r2)] (p2) {};
\end{scope}

\node[ptitle, below=6pt of p2.south west, anchor=north west, xshift=0.15cm] (t3) {(c) Prefix Contamination};
\node[mlabel, right=2pt of t3] {GEPA $\cdot$ GPT-OSS-20B $\cdot$ Step\,14};

\node[ctag, below=2.5pt of t3.south west, anchor=north west] (c3a) {\texttt{Start\_With} ``Google''};
\node[ctag, right=1.5pt of c3a] (c3b) {\texttt{No\_Commas}};
\node[ctag, right=1.5pt of c3b] (c3c) {\texttt{Length} 3 sent.};
\node[ctag, below=1.5pt of c3a.south west, anchor=north west] (c3d) {\texttt{End\_With} ``Teams''};
\node[ctag, right=1.5pt of c3d] (c3e) {\texttt{Topic} gratitude};
\node[ctag, right=1.5pt of c3e] (c3f) {\texttt{Tone} motivational};
\node[ctag, below=1.5pt of c3d.south west, anchor=north west] (c3g) {\texttt{Emotion} determination};

\node[callout, below=2.5pt of c3g.south west, anchor=north west] (r3) {
\textbf{\color{red!70!black}Response} \hfill \tikz\node[badge]{0/8};\\[-1pt]
{\ttfamily \colorbox{yellow!35}{\strut Here is what you asked for:} \colorbox{red!20}{\strut I recommend} \colorbox{green!20}{Google} Classroom and Microsoft Teams as powerful classroom management tools\ldots}\\[0.5pt]
{\color{red!55!black}\itshape Constraint: ``Start with Google.'' Evolved prompt injects prefix; 19/50 samples affected. Base LLM: 4/8.}
};

\begin{scope}[on background layer]
\node[panel=purple!45, fit=(t3)(c3c)(c3g)(r3)] (p3) {};
\end{scope}

\end{tikzpicture}%
}
\caption{Examples of observed failures. Each panel shows a single step within a schedule: Blue: All active constraints at that step, Red: Callouts show the model output with the failure cause highlighted.}
\label{fig:failures}
\end{figure}

Table~\ref{tab:main_results} reports selected metrics for Llama-3.3-70B, GPT-OSS-120B, and Gemma-4-26B-A4B; Figure~\ref{fig:main_results} compares mean satisfaction across all five backbones. Results for other models and efficiency comparisons are in Appendix~\ref{app:perbackbone}. The central finding is that no adaptation method achieves a significant improvement over the no-adaptation baseline (Base LLM) on any metric. For mean satisfaction, a paired analysis across all 15 backbone$\times$schedule cells finds no evidence of superiority (all Holm-corrected one-sided $p_{\mathrm{sup}}\geq0.84$); bootstrap confidence intervals, paired permutation tests, TOST equivalence analyses, and their unit-of-analysis caveat are reported in Appendix~\ref{app:significance}. On GPT-OSS models, adaptation is actively harmful (up to $-0.176$ mean satisfaction). On Llama models, methods converge within noise of the baseline while consuming up to 1.5$\times$ the latency. This points to a structural misalignment with the proactive setting. These findings hold on the rule-only schedule, which uses purely deterministic validators and no LLM judge, confirming they are not artifacts of judge noise (Appendix~\ref{app:ruleonly}). Peak forgetting confirms the pattern: adaptation methods increase forgetting by up to 84\% on GPT-OSS models (ACE: 0.330 vs.\ base 0.179 on 120B) due to context accumulation---as prompt artifacts grow (187$\to$9K chars for ACE), earlier constraint signals are diluted in the context window, causing previously-satisfied types to regress (Figure~\ref{fig:failures}).
The Gemma-4-26B-A4B results provide a third model-family check (Table~\ref{tab:main_results}; complete results in Appendix~\ref{app:gemma}). The closest method, Dynamic Cheatsheet, changes mean satisfaction only from $0.677$ to $0.678$ while consuming 2.4$\times$ as many tokens; every other adaptation method has lower mean satisfaction than Base LLM. Thus, the central trend is not specific to the Llama or GPT-OSS families.


\begin{table}[t]
\centering
\begingroup
\footnotesize
\setlength{\tabcolsep}{2pt}
\begin{tabular}{@{}l ccccc@{}}
\toprule
Method & $\overline{\text{sat}}$ & Forg.$\downarrow$ & Coll.$\downarrow$ & Sw.$\uparrow$ & UF$\uparrow$ \\
\midrule
\multicolumn{6}{c}{\textit{Llama-3.3-70B}} \\[2pt]
Base LLM     & 0.595 & 0.235 & 0.048 & 0.295 & 0.642 \\
ICL          & 0.592 & \textbf{0.188} & \textbf{0.033} & \textbf{0.301} & 0.650 \\
Dyn.\ Cheatsheet & 0.600 & 0.204 & 0.048 & \underline{0.300} & \textbf{0.693} \\
ACE          & \underline{0.602} & 0.212 & 0.035 & 0.290 & \underline{0.683} \\
GEPA         & \textbf{0.603} & \underline{0.198} & 0.048 & 0.291 & 0.643 \\
MIPROv2      & 0.595 & 0.209 & \underline{0.034} & 0.282 & 0.672 \\
\midrule
\multicolumn{6}{c}{\textit{GPT-OSS-120B}} \\[2pt]
Base LLM     & \textbf{0.630} & \textbf{0.179} & 0.053 & 0.309 & 0.674 \\
ICL          & 0.598 & \underline{0.211} & \textbf{0.040} & \textbf{0.317} & \textbf{0.730} \\
Dyn.\ Cheatsheet & \underline{0.610} & 0.224 & 0.050 & \underline{0.315} & \underline{0.709} \\
ACE          & 0.454 & 0.330 & 0.050 & 0.289 & 0.610 \\
GEPA         & 0.506 & 0.268 & \underline{0.049} & 0.301 & 0.648 \\
MIPROv2      & 0.571 & 0.222 & 0.053 & 0.256 & 0.635 \\
\midrule
\multicolumn{6}{c}{\textit{Gemma-4-26B-A4B}} \\[2pt]
Base LLM     & \underline{0.677} & 0.163 & 0.044 & \underline{0.325} & 0.702 \\
ICL          & 0.627 & 0.167 & 0.049 & 0.321 & 0.701 \\
Dyn.\ Cheatsheet & \textbf{0.678} & \textbf{0.134} & \textbf{0.031} & \textbf{0.334} & 0.717 \\
ACE          & 0.504 & 0.272 & 0.056 & 0.288 & 0.719 \\
GEPA         & 0.658 & \underline{0.161} & \underline{0.036} & 0.311 & \underline{0.749} \\
MIPROv2      & 0.660 & 0.172 & 0.054 & 0.312 & \textbf{0.767} \\
\bottomrule
\end{tabular}
\endgroup
\caption{Results on RECAP. $\overline{\text{sat}}$: mean satisfaction; Forg.: peak forgetting; Coll.: collateral damage; Sw.: edit switch; UF: unlearning fidelity. Bold/underline: best/second-best within each backbone. Full results with std. dev. are in Appendices~\ref{app:perbackbone} and~\ref{app:gemma}.}
\label{tab:main_results}
\end{table}

\noindent\textbf{Scale dominates method.} Model size determines satisfaction scores more than the adaptation strategy. Llama-70B achieves ${\sim}0.60$ regardless of method (all $6$ fall within $0.592$--$0.603$), while Llama-8B plateaus near $0.49$ (Figure~\ref{fig:main_results}). The best method on 70B only shows negligible gains over Base LLM (GEPA, +0.008), yet costs 1.5$\times$ the latency (189s vs.\ 126s per step).

\noindent\textbf{Self-play fitness does not transfer.} The three optimization methods report high self-play pass rates during adaptation, yet actual test satisfaction shows no gain over Base LLM. Self-play optimization for the newly added constraint on a fixed synthetic task is not robust to regression on existing constraints embedded in the evaluation prompts.
The cost is substantial: GEPA's 17 extra calls per step add 64s of latency on 70B for no clear performance gain.

\noindent\textbf{More self-play compute does not help.} Scaling the self-play budget for ACE to 3$\times$ and 5$\times$ \emph{degrades} quality further: satisfaction drops from $0.454$ to $0.340$ as the playbook grows faster and the model's refusal rates rise $4$ times (Appendix~\ref{app:selfplay_ablation}).
The failure is structural, not computational.

\noindent\textbf{Closing the information gap yields expensive neutrality.} Providing the full constraint set to \texttt{adapt()} (not just the delta) recovers most of the deficit on GPT-OSS-120B ($0.454 \to 0.614$, vs.\ base $0.630$), but at 2.8$\times$ the token cost for near zero net improvement over Base LLM  (Appendix~\ref{app:full_constraints}).

\noindent\textbf{Accumulated artifacts introduce spurious contradictions.} Adaptive methods accumulate stale values and may introduce tighter constraints than specified (e.g., adding a sentence-count restriction when optimizing for word-count). When a new constraint conflicts with these outdated or hallucinated values, the LLM mistakenly refuses to generate. On GPT-OSS-120B, ACE's playbook grows to 9K characters over 20 steps, producing a 14\% refusal rate. MIPROv2's compiled prompt locks early values and is never revised, causing 31\% of responses on GPT-OSS-20B to follow stale directives. Llama models are more robust, treating system prompts as guidance rather than hard directives.

\noindent\textbf{Failure mode taxonomy.} We identify $6$ systematic failure modes that show structural limitations of the adaptation methods when tested under a proactive protocol (Table~\ref{tab:failures} in Appendix~\ref{app:failures}); Figure~\ref{fig:failures} shows $3$ of them as examples: (a) A locked prompt causes the model to follow stale directives from earlier steps, (b) Stale context triggers refusal on non-conflicting constraints, and (c) An evolved prompt injects a prefix that violates formatting constraints.

\FloatBarrier
\section{Conclusion}
\label{sec:conclusion}
We formalize proactive adaptation -- how agents handle evolving constraints without feedback or history, and introduce \methodname to evaluate it. Across 90 conditions, no adaptation method achieves a statistically significant aggregate improvement over a baseline that simply appends active constraints to the prompt, despite incurring higher costs. Because real-world constraints are often heterogeneous and non-enumerable, this gap highlights the urgent need for building efficient, regression-free adaptation methods for production agentic systems.


\section*{Limitations}
We note three limitations. First, all methods are prompt-level (no fine-tuning). Adaptation methods that explore model training specifically for this challenging continual learning setup is one potential future direction for progress in the future. Second, a comparison with reactive methods along with real-time feedback from a specialized oracle would further quantify the proactive gap. We encourage this experiment for future work, noting that self-play evidence presented in this work (near-perfect internal fitness vs.\ flat test performance) already demonstrates the gap is binding. Third, our experiments are based on task-agnostic instruction-following constraints derived from RECAST. While this is informative for general settings, niche task-specific settings may require different operation semantics, validators, and adaptation interfaces, and may exhibit different regression dynamics. Future work should therefore extend \methodname to domain-specific streams.

\section*{Ethical considerations}
Our work was approved by the established internal review procedure. We carefully verified the licensing information associated with all the datasets and LLMs used in this work, ensuring that their use was within their intended scope.

Our benchmark uses publicly available instruction-following data (Tulu~3 Persona IF prompts) that does not contain personal, sensitive, or harmful content. No human subjects were involved; all evaluation is automated. We note that proactive adaptation methods, if made effective in future work, could be misused to silently inject constraints that alter system behavior without user awareness (e.g., suppressing certain topics). We believe that benchmarking and understanding these methods transparently is a necessary step toward responsible deployment and appropriate safeguards.

\bibliography{references}

\appendix

\section{Protocol Pseudocode}
\label{app:protocol}

\begin{algorithm}[ht]
\caption{\methodname Evaluation Protocol}
\label{alg:protocol}
\footnotesize
\begin{algorithmic}[1]
\Require Schedule $\{(k, \text{op}_k)\}_{k=0}^{K}$, method $\mathcal{M}$, model $\theta$
\State $\mathcal{M}.\text{reset}()$
\For{$k = 0 \ldots K$}
    \State $\text{ctx} \leftarrow \textsc{StepContext}(\text{op}_k,\ \text{reg\_meta})$ \Comment{see App.~\ref{app:stepcontext}}
    \State $\mathcal{M}.\text{adapt}(\theta, \text{ctx})$ \Comment{proactive; no test results}
    \State Evaluate on all $\mathcal{C}_{0:k}$ + shadow-eval edited/deleted
\EndFor
\end{algorithmic}
\end{algorithm}

Here $\theta$ denotes the frozen LLM backbone, \texttt{reg\_meta} is the regression metadata (which types are new, retained, edited, or deleted), and \textsc{StepContext} bundles the operation and metadata into the adaptation interface.

\section{Constraint Taxonomy}
\label{app:taxonomy}

Table~\ref{tab:full_taxonomy} lists all 21 constraint types.

\begin{table*}[t]
\centering
\small
\setlength{\tabcolsep}{3pt}
\caption{Full constraint taxonomy. Rule-based types use deterministic validators; LLM-based types use an LLM judge.}
\label{tab:full_taxonomy}
\begin{tabular}{@{}lp{5.5cm}p{5.5cm}@{}}
\toprule
Type & Eval Method & Example \\
\midrule
\texttt{Length} & Rule: \texttt{evaluate\_word\_length} / \texttt{evaluate\_sentence\_length} & ``Keep your answer around 200 words'' \\
\texttt{Keyword} & Rule: \texttt{evaluate\_keyword} & ``Include `sustainability' 3 times'' \\
\texttt{Format} & Rule: \texttt{evaluate\_format} & ``Respond in JSON format'' \\
\texttt{Start\_With} & Rule: \texttt{evaluate\_start\_with} & ``Begin with the word `Dear'\,'' \\
\texttt{End\_With} & Rule: \texttt{evaluate\_end\_with} & ``End with `Sincerely'\,'' \\
\texttt{All\_Upper} & Rule: \texttt{check\_english\_uppercase} & ``Write entirely in uppercase'' \\
\texttt{All\_Lower} & Rule: \texttt{check\_english\_lowercase} & ``Write entirely in lowercase'' \\
\texttt{No\_Commas} & Rule: \texttt{contains\_no\_punctuation} & ``Do not use any commas'' \\
\midrule
\texttt{Style} & LLM Judge & ``Write in a formal academic style'' \\
\texttt{Topic} & LLM Judge & ``Focus on renewable energy'' \\
\texttt{Tone} & LLM Judge & ``Use an encouraging tone'' \\
\texttt{Emotion} & LLM Judge & ``Express gratitude'' \\
\texttt{Example} & LLM Judge & ``Include a real-world example'' \\
\texttt{Factuality} & LLM Judge & ``Ensure all facts are accurate'' \\
\texttt{Helpfulness} & LLM Judge & ``Provide actionable advice'' \\
\texttt{Language} & LLM Judge & ``Write in Spanish'' \\
\texttt{Numerical Constraints} & LLM Judge & ``List exactly 5 items'' \\
\texttt{Role Playing} & LLM Judge & ``Respond as a financial advisor'' \\
\texttt{Situation} & LLM Judge & ``Assume the reader is a beginner'' \\
\texttt{Background Info} & LLM Judge & ``Reference the 2024 climate report'' \\
\midrule
\texttt{Unclassified} & LLM Judge & Constraints not fitting above categories \\
\bottomrule
\end{tabular}
\end{table*}

\section{Schedule Configuration}
\label{app:config}

\begin{table}[ht]
\centering
\footnotesize
\setlength{\tabcolsep}{2pt}
\caption{The three evaluation schedules. A/E/D = adds/edits/deletes.}
\label{tab:schedules}
\begin{tabular}{@{}lcccl@{}}
\toprule
Schedule & Steps & A/E/D & Types & Rationale \\
\midrule
Interleaved-20 & 20 & 11/5/4 & 11 & Mixed ops \\
Clustered-20 & 20 & 11/5/4 & 11 & Phased A$\to$E$\to$D \\
Rule-Only-15 & 15 & 7/4/4 & 6 & Determ.\ eval \\
\bottomrule
\end{tabular}
\end{table}

\paragraph{Interleaved-20.}
Edits and deletes interleaved with adds throughout 20 steps. Types span rule-based and LLM-judged. Features: edit-on-edit (Length edited at steps~5 and~16), delete-after-edit (Style edited then deleted), variable post-deletion windows (12, 8, 5, 0 steps).

\paragraph{Clustered-20.}
Identical operations but phased: ADDs first (steps~0--10), then EDITs (steps~11--15), then DELETEs (steps~16--19). Clustered produces 12\% higher forgetting than interleaved (0.251 vs.\ 0.223), consistent with classical CL findings.

\paragraph{Rule-Only-15.}
Only 6 rule-based types with deterministic validators. This confirms that the findings are not artifacts of LLM-judge noise.

\begin{table}[ht]
\centering
\small
\begin{tabular}{cll}
\toprule
\textbf{Step ID} & \textbf{Operation} & \textbf{Type} \\
\midrule
0  & add    & Length      \\
1  & add    & Keyword     \\
2  & add    & Format      \\
3  & add    & Style       \\
4  & add    & Topic       \\
5  & edit   & Length      \\
6  & add    & Start\_With \\
7  & delete & Format      \\
8  & add    & End\_With   \\
9  & edit   & Style       \\
10 & add    & Tone        \\
11 & delete & Keyword     \\
12 & edit   & Topic       \\
13 & add    & Emotion     \\
14 & delete & Style       \\
15 & add    & Helpfulness \\
16 & edit   & Length      \\
17 & add    & Factuality  \\
18 & edit   & Tone        \\
19 & delete & End\_With   \\
\bottomrule
\end{tabular}
\caption{Interleaved-20 schedule configuration.}
\label{lst:config}
\end{table}


\section{Detailed Metric Definitions}
\label{app:metrics}

\paragraph{Satisfaction.}
For type~$t$ at step~$k$:
\begin{equation*}
\satrate(t,k) = \frac{|\{x : t \text{ satisfied}\}|}{|\{x : t \text{ active}\}|}
\end{equation*}
We report mean satisfaction across types and steps.

\paragraph{Peak forgetting.}
\begin{equation*}
\peakforg(t, k) = \max_{j < k} \satrate(t, j) - \satrate(t, k)
\end{equation*}
Clamped to $[0, \infty)$.

\paragraph{Collateral damage.}
When step~$k$ targets type~$t^*$, let $\mathcal{R}_k = \mathcal{C}_{0:k} \setminus \{t^*\}$:
\begin{equation*}
\text{coll}(k) = \frac{1}{|\mathcal{R}_k|} \!\sum_{t \in \mathcal{R}_k}\! \max\bigl(0,\; \Delta_t(k)\bigr)
\end{equation*}
where $\Delta_t(k) = \satrate(t, k{-}1) - \satrate(t, k)$.

\paragraph{Edit switch.}
\begin{equation*}
\text{switch} = \frac{\text{Adapted}}{\text{Both} + \text{Stale} + \text{Adapted}}
\end{equation*}
where Adapted = satisfies new but not old, Both = satisfies both, Stale = satisfies old but not new.

\paragraph{Unlearning Fidelity.}
When type~$d$ is deleted at step~$k$:
\begin{equation*}
\uf(d, k) = 1 - \frac{|r_\text{post}(d,k) - r_\text{def}(d)|}{\max\bigl(|r_\text{pre}(d) - r_\text{def}(d)|,\, \epsilon\bigr)}
\end{equation*}
where $r_\text{pre}$ is frozen at deletion, $r_\text{post}$ is shadow-evaluated at step~$k$, $r_\text{def}$ is the unconstrained default rate, and $\epsilon{=}0.05$.
Clamped to $[0,1]$. Three variants: immediate, sustained (reported), rebound.

\paragraph{Trajectory metrics.}
$\overline{\text{sat}}$ = mean across steps.
Init.\ = satisfaction at first introduction (forward transfer).
Final = last-step average.
BWT = mean change between first appearance and final step per type.

\paragraph{Efficiency.}
Step latency = adapt time + generation time (seconds). Tokens = total LLM tokens consumed across adaptation and inference.

\section{Statistical Significance of the Main Claim}
\label{app:significance}

Within each backbone$\times$schedule cell, every method uses the same base
prompts, constraint stream, and random seed. For each method, we analyze the
$5\times3=15$ paired cell differences
$\Delta\overline{\text{sat}}=\overline{\text{sat}}_{\text{method}}-\overline{\text{sat}}_{\text{Base}}$
(positive favors adaptation). We report percentile-bootstrap $95\%$ CIs
($10^4$ resamples), exact paired sign-flip tests ($2^{15}$ enumerations;
two-sided $p_{\mathrm{perm}}$ and one-sided superiority $p_{\mathrm{sup}}$),
both Holm--Bonferroni corrected across five methods, and TOST equivalence at a
$\pm0.05$ margin ($90\%$ CI within the margin; $\alpha{=}0.05$). These
nonparametric procedures assume exchangeable paired units; all randomized
analyses use seed~42.

\begin{table*}[t]
\centering\small
\setlength{\tabcolsep}{4pt}
\caption{Paired mean-satisfaction results. Cell-level statistics use $n{=}15$
backbone$\times$schedule cells; $p$-values are exact sign-flip tests with Holm
correction, and ``Equiv.'' denotes TOST equivalence to Base within $\pm0.05$.
``Wins'' counts positive cells. $\Delta_{\mathrm{RO}}$ is the pooled
prompt-macro difference on deterministic Rule-Only-15 ($n{=}250$ pooled
backbone--prompt observations; $n{\approx}50$ prompts per backbone);
$^{\ast}$ marks a base-prompt-clustered $95\%$ bootstrap CI entirely below zero.}
\label{tab:paired_satisfaction}
\begin{tabular}{@{}l r c c c c c c@{}}
\toprule
Method & $\Delta\overline{\text{sat}}$ & 95\% CI & $p_{\text{perm}}$ & $p_{\text{sup}}$ & Wins & Equiv. & $\Delta_{\mathrm{RO}}$ \\
\midrule
ICL              & $-0.038$ & $[-0.058,-0.022]$ & $<0.001$ & $1.00$ & $1/15$ & No  & $-0.018^{\ast}$ \\
Dyn.\ Cheatsheet & $-0.006$ & $[-0.017,+0.004]$ & $0.310$  & $0.84$ & $8/15$ & Yes & $-0.004$ \\
ACE              & $-0.117$ & $[-0.179,-0.059]$ & $0.004$  & $1.00$ & $4/15$ & No  & $-0.032^{\ast}$ \\
GEPA             & $-0.046$ & $[-0.085,-0.015]$ & $0.026$  & $1.00$ & $5/15$ & No  & $-0.025^{\ast}$ \\
MIPROv2          & $-0.040$ & $[-0.083,-0.003]$ & $0.182$  & $0.95$ & $8/15$ & No  & $-0.060^{\ast}$ \\
\bottomrule
\end{tabular}
\end{table*}

All cell-level estimates are non-positive, all superiority tests satisfy
$p_{\mathrm{sup}}\geq0.84$, and the largest upper confidence bound is $+0.004$;
thus, every method excludes a $0.05$-point gain (Table~\ref{tab:paired_satisfaction}).
Dynamic Cheatsheet alone is equivalent to Base within $\pm0.05$. Holm-corrected
two-sided tests find ICL, ACE, and GEPA worse than Base; MIPROv2 is not
significant after correction.

The 15 crossed cells share backbones and schedules, so cell-level inference may
be anti-conservative. With backbones as the units, every mean effect remains
negative, but $n{=}5$ gives a minimum two-sided sign-flip $p$-value of $0.0625$.
The deterministic Rule-Only sensitivity check instead resamples complete
base-prompt clusters, retaining all five backbone observations per prompt. Its
pooled CIs are below zero for ICL, ACE, GEPA, and MIPROv2, while Dynamic
Cheatsheet includes zero (last column of Table~\ref{tab:paired_satisfaction});
the no-improvement result therefore does not depend on an LLM judge or on
treating repeated prompt observations as independent. This prompt-macro
estimator weights prompts equally and can differ slightly from the type/step-macro
$\overline{\text{sat}}$ in Table~\ref{tab:ruleonly}.

In the same cell-level analysis, no method improves peak forgetting, unlearning
fidelity, edit-switch, or collateral damage after Holm correction; the only
significant secondary effects are deteriorations by ACE (forgetting $+0.083$;
edit-switch $-0.025$). Cell-level efficiency contrasts are also positive for
every method: $+4.3$ to $+38.3$\,s per step and $+375{,}024$ to $+623{,}294$
tokens, with every corresponding $95\%$ bootstrap CI strictly above zero.

\section{Per-Backbone Results}
\label{app:perbackbone}

Table~\ref{tab:main_results} in the main body reports selected metrics for Llama-3.3-70B, GPT-OSS-120B, and Gemma-4-26B-A4B; Figure~\ref{fig:main_results} compares mean satisfaction across all five backbones. Tables~\ref{tab:llama318b}--\ref{tab:gptoss120b} report all metrics for the Llama and GPT-OSS model family backbones, and Table~\ref{tab:efficiency} consolidates their efficiency metrics. Complete Gemma-4-26B-A4B results are reported separately in Appendix~\ref{app:gemma}. All values are averaged over 3 schedules ($n{=}3$). Reported $\pm$ values are SDs across these heterogeneous schedules, not estimates of random-seed uncertainty.

\begin{table*}[ht]
\centering
\small
\caption{Trajectory-level results for \textbf{Llama 3.1 8B} (averaged over 3 schedules). Bold = best; underline = second best.}
\label{tab:llama318b}
\setlength{\tabcolsep}{4pt}
\begin{tabular}{@{}l ccc ccc cc@{}}
\toprule
& \multicolumn{3}{c}{Satisfaction} & \multicolumn{3}{c}{Regression} & \multicolumn{2}{c}{Edit / Delete} \\
\cmidrule(lr){2-4} \cmidrule(lr){5-7} \cmidrule(lr){8-9}
Method & $\overline{\text{sat}}$ & Init. & Final & Forg.$\downarrow$ & Coll.$\downarrow$ & BWT$\uparrow$ & Switch$\uparrow$ & $\text{UF}_{\text{sus}}\!\uparrow$ \\
\midrule
Base LLM         & \underline{0.490{\scriptsize$\pm$0.020}} & \textbf{0.665{\scriptsize$\pm$0.037}} & \underline{0.536{\scriptsize$\pm$0.036}} & \textbf{0.204{\scriptsize$\pm$0.032}} & 0.058{\scriptsize$\pm$0.021} & -0.096{\scriptsize$\pm$0.066} & 0.245{\scriptsize$\pm$0.116} & 0.589{\scriptsize$\pm$0.175} \\
ICL              & 0.399{\scriptsize$\pm$0.052} & 0.598{\scriptsize$\pm$0.097} & 0.302{\scriptsize$\pm$0.130} & 0.269{\scriptsize$\pm$0.055} & 0.058{\scriptsize$\pm$0.028} & -0.148{\scriptsize$\pm$0.044} & 0.252{\scriptsize$\pm$0.156} & \underline{0.623{\scriptsize$\pm$0.200}} \\
Dyn.\ Cheat.     & 0.480{\scriptsize$\pm$0.042} & \underline{0.652{\scriptsize$\pm$0.054}} & 0.497{\scriptsize$\pm$0.051} & 0.220{\scriptsize$\pm$0.078} & \underline{0.049{\scriptsize$\pm$0.021}} & \textbf{0.016{\scriptsize$\pm$0.152}} & 0.253{\scriptsize$\pm$0.133} & 0.607{\scriptsize$\pm$0.158} \\
ACE              & 0.425{\scriptsize$\pm$0.106} & 0.605{\scriptsize$\pm$0.108} & 0.425{\scriptsize$\pm$0.135} & 0.252{\scriptsize$\pm$0.106} & \textbf{0.043{\scriptsize$\pm$0.019}} & \underline{0.009{\scriptsize$\pm$0.029}} & 0.240{\scriptsize$\pm$0.147} & 0.565{\scriptsize$\pm$0.166} \\
GEPA             & 0.483{\scriptsize$\pm$0.034} & 0.573{\scriptsize$\pm$0.116} & 0.492{\scriptsize$\pm$0.039} & \underline{0.209{\scriptsize$\pm$0.045}} & 0.050{\scriptsize$\pm$0.020} & -0.146{\scriptsize$\pm$0.171} & \underline{0.253{\scriptsize$\pm$0.126}} & \textbf{0.639{\scriptsize$\pm$0.190}} \\
MIPROv2          & \textbf{0.507{\scriptsize$\pm$0.021}} & 0.647{\scriptsize$\pm$0.081} & \textbf{0.550{\scriptsize$\pm$0.031}} & 0.212{\scriptsize$\pm$0.013} & 0.052{\scriptsize$\pm$0.019} & -0.197{\scriptsize$\pm$0.113} & \textbf{0.257{\scriptsize$\pm$0.145}} & 0.596{\scriptsize$\pm$0.158} \\
\bottomrule
\end{tabular}
\end{table*}

\begin{table*}[ht]
\centering
\small
\caption{Trajectory-level results for \textbf{GPT-OSS 20B} (averaged over 3 schedules). Bold = best; underline = second best.}
\label{tab:gptoss20b}
\setlength{\tabcolsep}{4pt}
\begin{tabular}{@{}l ccc ccc cc@{}}
\toprule
& \multicolumn{3}{c}{Satisfaction} & \multicolumn{3}{c}{Regression} & \multicolumn{2}{c}{Edit / Delete} \\
\cmidrule(lr){2-4} \cmidrule(lr){5-7} \cmidrule(lr){8-9}
Method & $\overline{\text{sat}}$ & Init. & Final & Forg.$\downarrow$ & Coll.$\downarrow$ & BWT$\uparrow$ & Switch$\uparrow$ & $\text{UF}_{\text{sus}}\!\uparrow$ \\
\midrule
Base LLM         & \textbf{0.555{\scriptsize$\pm$0.016}} & 0.625{\scriptsize$\pm$0.123} & \textbf{0.518{\scriptsize$\pm$0.045}} & \textbf{0.184{\scriptsize$\pm$0.023}} & 0.056{\scriptsize$\pm$0.045} & \textbf{-0.153{\scriptsize$\pm$0.252}} & \underline{0.310{\scriptsize$\pm$0.170}} & 0.611{\scriptsize$\pm$0.174} \\
ICL              & 0.539{\scriptsize$\pm$0.022} & \textbf{0.754{\scriptsize$\pm$0.026}} & 0.486{\scriptsize$\pm$0.093} & \underline{0.221{\scriptsize$\pm$0.009}} & \underline{0.055{\scriptsize$\pm$0.021}} & -0.172{\scriptsize$\pm$0.194} & 0.298{\scriptsize$\pm$0.166} & \underline{0.644{\scriptsize$\pm$0.161}} \\
Dyn.\ Cheat.     & \underline{0.549{\scriptsize$\pm$0.036}} & \underline{0.672{\scriptsize$\pm$0.035}} & 0.480{\scriptsize$\pm$0.112} & 0.248{\scriptsize$\pm$0.067} & \textbf{0.049{\scriptsize$\pm$0.016}} & -0.274{\scriptsize$\pm$0.148} & \textbf{0.317{\scriptsize$\pm$0.209}} & 0.612{\scriptsize$\pm$0.173} \\
ACE              & 0.379{\scriptsize$\pm$0.106} & 0.627{\scriptsize$\pm$0.065} & 0.277{\scriptsize$\pm$0.120} & 0.315{\scriptsize$\pm$0.110} & 0.065{\scriptsize$\pm$0.030} & \underline{-0.161{\scriptsize$\pm$0.177}} & 0.251{\scriptsize$\pm$0.188} & 0.559{\scriptsize$\pm$0.253} \\
GEPA             & 0.466{\scriptsize$\pm$0.053} & 0.658{\scriptsize$\pm$0.063} & \underline{0.496{\scriptsize$\pm$0.165}} & 0.284{\scriptsize$\pm$0.018} & 0.071{\scriptsize$\pm$0.034} & -0.213{\scriptsize$\pm$0.095} & 0.283{\scriptsize$\pm$0.188} & 0.635{\scriptsize$\pm$0.144} \\
MIPROv2          & 0.415{\scriptsize$\pm$0.068} & 0.662{\scriptsize$\pm$0.056} & 0.340{\scriptsize$\pm$0.146} & 0.281{\scriptsize$\pm$0.044} & 0.062{\scriptsize$\pm$0.012} & -0.237{\scriptsize$\pm$0.200} & 0.275{\scriptsize$\pm$0.168} & \textbf{0.752{\scriptsize$\pm$0.248}} \\
\bottomrule
\end{tabular}
\end{table*}

\begin{table*}[ht]
\centering
\small
\caption{Trajectory-level results for \textbf{Llama 3.3 70B} (averaged over 3 schedules). Bold = best; underline = second best.}
\label{tab:llama3370b}
\setlength{\tabcolsep}{4pt}
\begin{tabular}{@{}l ccc ccc cc@{}}
\toprule
& \multicolumn{3}{c}{Satisfaction} & \multicolumn{3}{c}{Regression} & \multicolumn{2}{c}{Edit / Delete} \\
\cmidrule(lr){2-4} \cmidrule(lr){5-7} \cmidrule(lr){8-9}
Method & $\overline{\text{sat}}$ & Init. & Final & Forg.$\downarrow$ & Coll.$\downarrow$ & BWT$\uparrow$ & Switch$\uparrow$ & $\text{UF}_{\text{sus}}\!\uparrow$ \\
\midrule
Base LLM         & 0.595{\scriptsize$\pm$0.011} & \underline{0.653{\scriptsize$\pm$0.042}} & 0.638{\scriptsize$\pm$0.013} & 0.235{\scriptsize$\pm$0.030} & 0.048{\scriptsize$\pm$0.025} & -0.239{\scriptsize$\pm$0.107} & 0.295{\scriptsize$\pm$0.190} & 0.642{\scriptsize$\pm$0.154} \\
ICL              & 0.592{\scriptsize$\pm$0.016} & 0.586{\scriptsize$\pm$0.022} & 0.641{\scriptsize$\pm$0.022} & \textbf{0.188{\scriptsize$\pm$0.026}} & \textbf{0.033{\scriptsize$\pm$0.012}} & \underline{-0.015{\scriptsize$\pm$0.113}} & \textbf{0.301{\scriptsize$\pm$0.190}} & 0.650{\scriptsize$\pm$0.150} \\
Dyn.\ Cheat.     & 0.600{\scriptsize$\pm$0.010} & 0.595{\scriptsize$\pm$0.007} & 0.651{\scriptsize$\pm$0.028} & 0.204{\scriptsize$\pm$0.037} & 0.048{\scriptsize$\pm$0.034} & -0.029{\scriptsize$\pm$0.121} & \underline{0.300{\scriptsize$\pm$0.170}} & \textbf{0.693{\scriptsize$\pm$0.089}} \\
ACE              & \underline{0.602{\scriptsize$\pm$0.014}} & 0.629{\scriptsize$\pm$0.040} & \textbf{0.668{\scriptsize$\pm$0.049}} & 0.212{\scriptsize$\pm$0.005} & 0.035{\scriptsize$\pm$0.021} & -0.073{\scriptsize$\pm$0.070} & 0.290{\scriptsize$\pm$0.176} & \underline{0.683{\scriptsize$\pm$0.111}} \\
GEPA             & \textbf{0.603{\scriptsize$\pm$0.004}} & \textbf{0.671{\scriptsize$\pm$0.100}} & \underline{0.668{\scriptsize$\pm$0.023}} & \underline{0.198{\scriptsize$\pm$0.006}} & 0.048{\scriptsize$\pm$0.026} & \textbf{-0.013{\scriptsize$\pm$0.181}} & 0.291{\scriptsize$\pm$0.157} & 0.643{\scriptsize$\pm$0.176} \\
MIPROv2          & 0.595{\scriptsize$\pm$0.009} & 0.618{\scriptsize$\pm$0.029} & 0.634{\scriptsize$\pm$0.042} & 0.209{\scriptsize$\pm$0.033} & \underline{0.034{\scriptsize$\pm$0.012}} & -0.218{\scriptsize$\pm$0.094} & 0.282{\scriptsize$\pm$0.175} & 0.672{\scriptsize$\pm$0.134} \\
\bottomrule
\end{tabular}
\end{table*}

\begin{table*}[ht]
\centering
\small
\caption{Trajectory-level results for \textbf{GPT-OSS 120B} (averaged over 3 schedules). Bold = best; underline = second best.}
\label{tab:gptoss120b}
\setlength{\tabcolsep}{4pt}
\begin{tabular}{@{}l ccc ccc cc@{}}
\toprule
& \multicolumn{3}{c}{Satisfaction} & \multicolumn{3}{c}{Regression} & \multicolumn{2}{c}{Edit / Delete} \\
\cmidrule(lr){2-4} \cmidrule(lr){5-7} \cmidrule(lr){8-9}
Method & $\overline{\text{sat}}$ & Init. & Final & Forg.$\downarrow$ & Coll.$\downarrow$ & BWT$\uparrow$ & Switch$\uparrow$ & $\text{UF}_{\text{sus}}\!\uparrow$ \\
\midrule
Base LLM         & \textbf{0.630{\scriptsize$\pm$0.023}} & \textbf{0.683{\scriptsize$\pm$0.078}} & \textbf{0.639{\scriptsize$\pm$0.031}} & \textbf{0.179{\scriptsize$\pm$0.021}} & 0.053{\scriptsize$\pm$0.027} & \underline{0.033{\scriptsize$\pm$0.131}} & 0.309{\scriptsize$\pm$0.192} & 0.674{\scriptsize$\pm$0.121} \\
ICL              & 0.598{\scriptsize$\pm$0.023} & \underline{0.682{\scriptsize$\pm$0.043}} & 0.575{\scriptsize$\pm$0.056} & \underline{0.211{\scriptsize$\pm$0.015}} & \textbf{0.040{\scriptsize$\pm$0.013}} & \textbf{0.075{\scriptsize$\pm$0.023}} & \textbf{0.317{\scriptsize$\pm$0.207}} & \textbf{0.730{\scriptsize$\pm$0.121}} \\
Dyn.\ Cheat.     & \underline{0.610{\scriptsize$\pm$0.036}} & 0.624{\scriptsize$\pm$0.060} & 0.587{\scriptsize$\pm$0.054} & 0.224{\scriptsize$\pm$0.034} & 0.050{\scriptsize$\pm$0.022} & -0.094{\scriptsize$\pm$0.179} & \underline{0.315{\scriptsize$\pm$0.212}} & \underline{0.709{\scriptsize$\pm$0.108}} \\
ACE              & 0.454{\scriptsize$\pm$0.151} & 0.639{\scriptsize$\pm$0.078} & 0.351{\scriptsize$\pm$0.222} & 0.330{\scriptsize$\pm$0.095} & 0.050{\scriptsize$\pm$0.014} & -0.180{\scriptsize$\pm$0.170} & 0.289{\scriptsize$\pm$0.212} & 0.610{\scriptsize$\pm$0.198} \\
GEPA             & 0.506{\scriptsize$\pm$0.100} & 0.663{\scriptsize$\pm$0.177} & 0.524{\scriptsize$\pm$0.153} & 0.268{\scriptsize$\pm$0.072} & \underline{0.049{\scriptsize$\pm$0.028}} & -0.059{\scriptsize$\pm$0.119} & 0.301{\scriptsize$\pm$0.197} & 0.648{\scriptsize$\pm$0.145} \\
MIPROv2          & 0.571{\scriptsize$\pm$0.113} & 0.629{\scriptsize$\pm$0.025} & \underline{0.620{\scriptsize$\pm$0.025}} & 0.222{\scriptsize$\pm$0.082} & 0.053{\scriptsize$\pm$0.014} & -0.030{\scriptsize$\pm$0.216} & 0.256{\scriptsize$\pm$0.098} & 0.635{\scriptsize$\pm$0.054} \\
\bottomrule
\end{tabular}
\end{table*}

\begin{table*}[ht]
\centering
\small
\caption{Efficiency metrics across the four original backbones (averaged over 3 schedules). Step latency includes adaptation and generation time (seconds). Tokens is the total LLM tokens consumed per run. Gemma efficiency is reported separately in Appendix~\ref{app:gemma}.}
\label{tab:efficiency}
\setlength{\tabcolsep}{4pt}
\begin{tabular}{@{}l cc cc cc cc@{}}
\toprule
& \multicolumn{2}{c}{Llama 3.1 8B} & \multicolumn{2}{c}{GPT-OSS 20B} & \multicolumn{2}{c}{Llama 3.3 70B} & \multicolumn{2}{c}{GPT-OSS 120B} \\
\cmidrule(lr){2-3} \cmidrule(lr){4-5} \cmidrule(lr){6-7} \cmidrule(lr){8-9}
Method & Lat.(s) & Tok. & Lat.(s) & Tok. & Lat.(s) & Tok. & Lat.(s) & Tok. \\
\midrule
Base LLM     & 108.6 & 491K & 61.3 & 643K & 125.5 & 468K & 97.2 & 796K \\
ICL          & 139.5 & 1,122K & 68.2 & 1,045K & 131.8 & 935K & 98.4 & 1,103K \\
Dyn.\ Cheat. & 114.5 & 918K & 72.8 & 1,223K & 128.7 & 767K & 95.0 & 1,351K \\
ACE          & 129.7 & 1,035K & 89.1 & 1,578K & 136.0 & 819K & 109.7 & 1,623K \\
GEPA         & 141.0 & 852K & 106.0 & 1,362K & 189.4 & 841K & 135.5 & 1,459K \\
MIPROv2      & 127.3 & 764K & 74.4 & 861K & 136.0 & 719K & 130.5 & 1,660K \\
\bottomrule
\end{tabular}
\end{table*}

\section{Rule-Only Schedule Analysis}
\label{app:ruleonly}

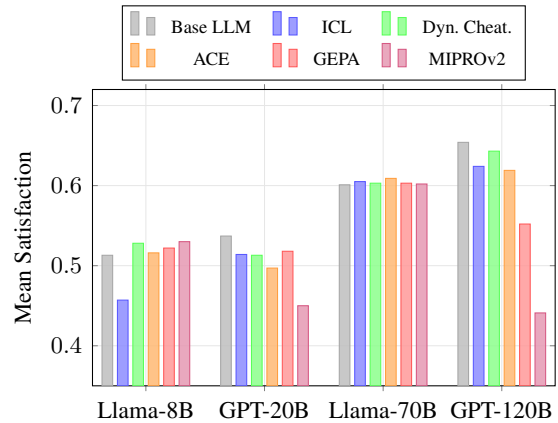
\begin{figure}[ht]
\centering
\begin{tikzpicture}
\begin{axis}[
    width=\linewidth,
    height=5.5cm,
    ybar=1.8pt,
    bar width=4pt,
    ylabel={Mean Satisfaction},
    ylabel style={font=\small},
    symbolic x coords={Llama-8B, GPT-20B, Llama-70B, GPT-120B},
    xtick=data,
    xticklabel style={font=\small},
    yticklabel style={font=\small},
    ymin=0.35, ymax=0.72,
    ytick={0.4, 0.5, 0.6, 0.7},
    legend style={
        font=\scriptsize,
        at={(0.5,1.03)},
        anchor=south,
        legend columns=3,
        column sep=4pt,
        /tikz/every even column/.append style={column sep=6pt},
    },
    every axis plot/.append style={fill opacity=0.85},
    grid=major,
    grid style={gray!20, line width=0.3pt},
    major tick length=2pt,
    enlarge x limits=0.15,
]

\addplot[fill=gray!50, draw=gray!70] coordinates {
    (Llama-8B, 0.513)
    (GPT-20B, 0.537)
    (Llama-70B, 0.601)
    (GPT-120B, 0.654)
};
\addplot[fill=blue!45, draw=blue!65] coordinates {
    (Llama-8B, 0.457)
    (GPT-20B, 0.514)
    (Llama-70B, 0.605)
    (GPT-120B, 0.624)
};
\addplot[fill=green!45, draw=green!65] coordinates {
    (Llama-8B, 0.528)
    (GPT-20B, 0.513)
    (Llama-70B, 0.603)
    (GPT-120B, 0.643)
};
\addplot[fill=orange!55, draw=orange!75] coordinates {
    (Llama-8B, 0.516)
    (GPT-20B, 0.497)
    (Llama-70B, 0.609)
    (GPT-120B, 0.619)
};
\addplot[fill=red!40, draw=red!65] coordinates {
    (Llama-8B, 0.522)
    (GPT-20B, 0.518)
    (Llama-70B, 0.603)
    (GPT-120B, 0.552)
};
\addplot[fill=purple!40, draw=purple!65] coordinates {
    (Llama-8B, 0.530)
    (GPT-20B, 0.450)
    (Llama-70B, 0.602)
    (GPT-120B, 0.441)
};

\legend{Base LLM, ICL, Dyn.\ Cheat., ACE, GEPA, MIPROv2}
\end{axis}
\end{tikzpicture}
\caption{Rule-Only-15 mean satisfaction (deterministic evaluation; no LLM judge). Base LLM leads on both GPT-OSS backbones, while all six conditions fall within 0.008 on Llama-70B.}
\label{fig:ruleonly_bar}
\end{figure}

The Rule-Only-15 schedule uses exclusively deterministic constraint types (Length, Keyword, Format, Start\_With, End\_With, No\_Commas), eliminating LLM judge noise entirely.
Figure~\ref{fig:ruleonly_bar} shows a similar pattern under deterministic evaluation across backbones: Base LLM leads on both GPT-OSS backbones, all methods fall within 0.008 on Llama-70B, and the largest adaptation gains on Llama-8B and Llama-70B are small ($+0.017$ and $+0.008$). Adaptation is especially harmful on GPT-OSS-120B (MIPROv2 drops to 0.441 and GEPA to 0.552, vs.\ 0.654 for Base). The prompt-clustered paired analysis, including Gemma-4-26B-A4B, is reported in Table~\ref{tab:paired_satisfaction}.
This reproduces the pooled no-improvement pattern without LLM-judge variability.

\begin{table}[ht]
\centering
\footnotesize
\setlength{\tabcolsep}{2pt}
\caption{Full results for the Rule-Only-15 schedule (deterministic evaluation only). Bold marks the best value within each backbone and metric.}
\label{tab:ruleonly}
\begin{tabular}{@{}ll ccccc@{}}
\toprule
BB & Method & $\overline{\text{sat}}$ & Forg.$\downarrow$ & Coll.$\downarrow$ & Sw.$\uparrow$ & UF$\uparrow$ \\
\midrule
\multirow{6}{*}{\rotatebox{90}{\scriptsize 8B}}
& Base LLM & 0.513 & 0.206 & 0.035 & 0.379 & 0.791 \\
& ICL & 0.457 & 0.222 & 0.036 & \textbf{0.431} & 0.851 \\
& Dyn.\ Ch. & 0.528 & 0.181 & \textbf{0.024} & 0.406 & 0.789 \\
& ACE & 0.516 & 0.198 & 0.026 & 0.410 & 0.750 \\
& GEPA & 0.522 & \textbf{0.162} & 0.027 & 0.398 & \textbf{0.857} \\
& MIPROv2 & \textbf{0.530} & 0.199 & 0.029 & 0.424 & 0.778 \\
\midrule
\multirow{6}{*}{\rotatebox{90}{\scriptsize 20B}}
& Base LLM & \textbf{0.537} & \textbf{0.159} & \textbf{0.025} & 0.506 & 0.812 \\
& ICL & 0.514 & 0.231 & 0.033 & 0.488 & 0.807 \\
& Dyn.\ Ch. & 0.513 & 0.219 & 0.031 & \textbf{0.557} & 0.811 \\
& ACE & 0.497 & 0.209 & 0.037 & 0.468 & \textbf{0.851} \\
& GEPA & 0.518 & 0.303 & 0.033 & 0.501 & 0.789 \\
& MIPROv2 & 0.450 & 0.254 & 0.049 & 0.467 & 0.751 \\
\midrule
\multirow{6}{*}{\rotatebox{90}{\scriptsize 70B}}
& Base LLM & 0.601 & 0.269 & 0.020 & 0.514 & 0.817 \\
& ICL & 0.605 & 0.212 & 0.019 & \textbf{0.519} & 0.819 \\
& Dyn.\ Ch. & 0.603 & 0.242 & 0.020 & 0.493 & 0.790 \\
& ACE & \textbf{0.609} & 0.212 & \textbf{0.017} & 0.493 & 0.810 \\
& GEPA & 0.603 & \textbf{0.191} & 0.018 & 0.470 & \textbf{0.839} \\
& MIPROv2 & 0.602 & 0.244 & 0.021 & 0.485 & 0.818 \\
\midrule
\multirow{6}{*}{\rotatebox{90}{\scriptsize 120B}}
& Base LLM & \textbf{0.654} & \textbf{0.175} & \textbf{0.022} & 0.531 & 0.794 \\
& ICL & 0.624 & 0.219 & 0.031 & 0.556 & 0.825 \\
& Dyn.\ Ch. & 0.643 & 0.199 & 0.025 & \textbf{0.560} & 0.823 \\
& ACE & 0.619 & 0.254 & 0.039 & 0.533 & \textbf{0.834} \\
& GEPA & 0.552 & 0.278 & 0.026 & 0.528 & 0.662 \\
& MIPROv2 & 0.441 & 0.316 & 0.037 & 0.368 & 0.584 \\
\bottomrule
\end{tabular}
\end{table}

\section{Failure Mode Analysis}
\label{app:failures}

\begin{table*}[t]
\centering
\small
\caption{Six failure modes observed across the 72 original production runs (four backbones $\times$ six methods $\times$ three schedules; Gemma runs are not included). Each represents a structural misalignment between the method's design assumptions and the proactive protocol. $\Delta$sat is relative to Base LLM on the worst-affected backbone.}
\label{tab:failures}
\setlength{\tabcolsep}{3pt}
\begin{tabular}{@{}llll p{5.8cm} r@{}}
\toprule
\# & Mode & Method & Backbone & Mechanism & $\Delta$sat \\
\midrule
1 & Spec.\ Lock & MIPROv2 & 20B & Compiled prompt hardcodes early constraint values (topic, prefix, keywords) and is never revised after edits. Model follows stale system prompt directives instead of current constraints. Model may even echo system prompt content. & $-0.140$ \\
2 & Refusal Cascade & ACE & 120B & Playbook grows to 9K chars; model perceives phantom contradictions between accumulated rules and constraints. 12.4\% average refusal rate. & $-0.176$ \\
3 & Demo Noise & ICL & 8B & Synthetic demos bloat context 90--130\%; surface patterns (preambles, punctuation) conflict with active constraints. & $-0.091$ \\
4 & Prefix Contam. & GEPA & 20B & Evolved prompt injects fixed prefix (``Here is what you asked for:'') violating all Start\_With constraints. 19\% of samples affected. & $-0.089$ \\
5 & Inert Growth & Dyn.\ Cheat. & 120B & Cheatsheet grows to 4K chars of generic advice; 62\% of responses identical to base. & $-0.020$ \\
6 & Anti-Unlearn & All & 120B & Topic-keyword semantic correlation causes deleted keywords to persist in 20--40\% of responses. & Keyword UF${\leq}0.62$ \\
\bottomrule
\end{tabular}
\end{table*}

We identify six systematic failure modes across the 72 original production runs.
Each illustrates a structural limitation of prompt-level continual adaptation under the proactive protocol.

\paragraph{Failure 1: Specification Lock (MIPROv2, GPT-OSS-20B).}
MIPROv2's compiled prompt hardcodes constraint values from early steps and never revises them after edits. Self-validation reports perfect scores despite a $-0.140$ gap versus base.
\begin{quote}\small
\textbf{Clustered Step 12, GPT-OSS-20B}: Current constraints require Start\_With ``To'', keywords ``Fair Trade Certification'' and ``transparent reports'', topic about action movies. Compiled prompt locks stale values: topic=DACA, prefix=``In addressing the legal challenges of DACA.'' Response follows the stale system prompt (1/12). Base LLM follows current constraints (5/12).
\end{quote}

\paragraph{Failure 2: Progressive Refusal Cascade (ACE, GPT-OSS-120B).}
ACE's playbook grows from 187 to 9,085 characters over 20 steps. The model perceives conflicts between accumulated rules and current constraints, with a 12.4\% refusal rate averaged across schedules.
\begin{quote}\small
\textbf{Step 15, GPT-OSS-120B}: ``I'm sorry, but I can't fulfill this request as it contains conflicting constraints that cannot be satisfied simultaneously.'' Base LLM satisfies 4/9 constraints---no actual conflict exists.
\end{quote}

\paragraph{Failure 3: Demo Noise (ICL, Llama-3.1-8B).}
Synthetic demonstrations bloat context by 90--130\% and introduce surface patterns (preambles, list punctuation) that conflict with active constraints.
\begin{quote}\small
\textbf{Step 7, Llama-8B}: Active constraints include Start\_With (``Google''), No\_Commas, Length ($\leq$3 sentences). ICL response begins ``Here are two classroom management tools\ldots'', uses commas, generates 107 words. Base LLM correctly starts ``Google Classroom is a game-changer\ldots'' using dashes instead of commas (5/8 vs.\ ICL 0/8).
\end{quote}

\paragraph{Failure 4: Prefix Contamination (GEPA, GPT-OSS-20B).}
GEPA's evolved prompts inject a fixed prefix (``Here is what you asked for:'') into 19/50 responses at affected steps, violating all Start\_With constraints.
\begin{quote}\small
\textbf{Step 14, GPT-OSS-20B}: Constraint ``Start with `Google'\,''. GEPA response starts ``Here is what you asked for: I recommend Google Classroom\ldots'' (0/8). Base correctly starts ``Google Classroom and Schoology are the tools\ldots'' (4/8).
\end{quote}

\paragraph{Failure 5: Inert Cheatsheet Growth (Dynamic Cheatsheet, GPT-OSS-120B).}
The cheatsheet grows to 3,945 characters of general strategies, yet 62\% of responses are identical to Base LLM. Its mean difference from Base across schedules is $-0.020$.
\begin{quote}\small
\textbf{Step 16, GPT-OSS-120B}: Cheatsheet advice ``Include testimonials for credibility'' introduces uppercase in a response requiring All\_Lower, causing the cheatsheet to \emph{harm} (5/8 vs.\ base 6/8).
\end{quote}

\paragraph{Failure 6: Keyword Anti-Unlearning (all methods, GPT-OSS-120B).}
After Keyword deletion, target keywords persist in 20--40\% of responses due to topic-keyword semantic correlation. Keyword-specific UF peaks at 0.62, never reaching 1.0.
\begin{quote}\small
\textbf{Step 14, GPT-OSS-120B}: Deleted keyword ``ethnographic'' appears naturally (``a recent ethnographic fieldwork project\ldots'') because it is semantically related to the topic. Structural constraints (End\_With) achieve near-perfect unlearning (3.7\% residual).
\end{quote}

\section{Per-Method Behavioral Profiles}
\label{app:methods}

\begin{table}[ht]
\centering
\small
\setlength{\tabcolsep}{3pt}
\caption{Method behavioral profiles. Delta and Win\% use all 15 backbone$\times$schedule cells; Win\% is the fraction with a positive paired difference from Base. Best and Worst BB use the mean paired difference for adaptation methods and mean satisfaction for Base. Adapt(s) is the mean total adaptation time per run measured over the 12 original cells.}
\label{tab:method_profile}
\begin{tabular}{@{}lccccc@{}}
\toprule
Method & Delta & Win\% & Best BB & Worst BB & Adapt(s) \\
\midrule
Base LLM & --- & --- & Gemma & 8B & 0 \\
ICL & $-0.038$ & 7\% & 70B & 8B & 24 \\
Dyn.\ Cheat. & $-0.006$ & 53\% & 70B & 120B & 64 \\
ACE & $-0.117$ & 27\% & 70B & 20B/120B & 224 \\
GEPA & $-0.046$ & 33\% & 70B & 120B & 739 \\
MIPROv2 & $-0.040$ & 53\% & 8B & 20B & 442 \\
\bottomrule
\end{tabular}
\end{table}

\textbf{Base LLM} succeeds by doing nothing harmful---constraint text in the user prompt is sufficient.
\textbf{ICL} wins only one of 15 cells; demos add 90--130\% token overhead and introduce conflicting patterns.
\textbf{Dynamic Cheatsheet} is the safest adaptive method (53\% nominal win rate, $-0.006$ mean delta), consistent with its use of general strategies rather than specific values; the cell-level TOST analysis finds it statistically equivalent to Base LLM rather than superior (Table~\ref{tab:paired_satisfaction}).
\textbf{ACE} is the most polarized: +0.007 on 70B but $-0.176$ on GPT-OSS-120B due to progressive refusal cascades.
\textbf{GEPA} spends 311 additional API calls per run for +0.008 on 70B and $-0.124$ on 120B.
\textbf{MIPROv2} has a $+0.017$ mean difference on Llama-8B but collapses on GPT-OSS-20B ($-0.140$) via specification lock.
In the analyzed failure cases, GPT-OSS backbones copy system-prompt patterns literally, while Llama backbones more often treat them as guidance.

\section{Implementation Details}
\label{app:stepcontext}

\paragraph{Adaptation interface:}
The adaptation interface encapsulates all information available to a method at each step:

{\small
\begin{description}[nosep,leftmargin=1.5cm,style=nextline,font=\normalfont\texttt]
\item[step\_id:] Integer step identifier.
\item[step\_ops:] Operations at this step, including concrete constraint text (e.g., \texttt{[\{"op": "edit", "type": "Length", "new\_value": "Keep under 500 words"\}]}).
\item[step\_meta:] Step metadata (schedule name, operation counts, type inventory).
\item[regression\_meta:] Which types are new, retained, removed, edited, plus cumulative deleted/edited sets.
\item[prev\_state:] Opaque state from previous adaptation call; \texttt{None} for step~0.
\item[num\_samples:] Window size (number of samples per evaluation step).
\end{description}
}

\subsection{LLM-judge Details}
\label{app:judge_protocol}

\paragraph{LLM-judge prompt and protocol.}
We follow RECAST's per-constraint binary evaluation protocol
\citep{guo2025recast} and use Claude Sonnet~4.5 as the judge. The
prompt below shows the field structure and decision rule used in our evaluation:

{\footnotesize
\begin{quote}
\ttfamily
Act as an objective evaluator. Decide whether the model response fully
satisfies the single constraint below. Evaluate only this constraint, not the
other requirements in the instruction. If the response is ambiguous, contains
an error, or lacks enough information to verify the constraint, answer No.\\[2pt]
Input instruction: \{instruction with active constraints\}\\
Model response: \{candidate response\}\\
Constraint: \{one constraint\}\\[2pt]
Return JSON only:\\
\{``Analysis'': ``\ldots'', ``Answer'': ``Yes'' or ``No''\}
\end{quote}
}

Each LLM-judged active or shadow constraint is scored in a separate judge call;
the judge receives no method identity and no responses from competing methods.
``Yes'' maps to 1 and ``No'' to 0. The strict rule above treats partial,
ambiguous, or unverifiable compliance as a failure. Rule-based constraint types
bypass the LLM judge and use their deterministic validators. Shadow constraints
are supplied only to the scorer after generation, so they can measure stale
behavior without leaking deleted or replaced requirements to the evaluated
model. We use the same prompt and judge model for every method,
backbone, and schedule.

\paragraph{Hyperparameters:}
\label{app:hyperparams}
ICL: $\text{num\_demos}{=}3$, constraint-matched selection, pool cap~30.
Dynamic Cheatsheet \citep{suzgun2025dynamic}: max 4K tokens.
ACE: 4K token playbook budget, 4-call self-play pipeline.
GEPA: population size~4, 2~generations, elite size~2, 2K token budget (17 calls/step).
MIPROv2: 3~proposals/step, max 20~history entries (${\sim}11$ calls/step).
All self-play methods use the backbone as both generator and judge (2 calls per evaluation).

\paragraph{Efficiency metrics:}
\label{app:efficiency}
We track per-step: adaptation wall-clock time, generation wall-clock time, constraint scoring time, tokens consumed during adaptation, tokens for test responses, LLM API calls during adaptation, and LLM API calls for test generation.

\section{Self-Play Budget Ablation}
\label{app:selfplay_ablation}

We test whether ACE's underperformance is due to insufficient optimization by scaling the self-play budget from 1$\times$ (4~LLM calls/step; default) to 3$\times$ (12~calls, 3~Gen$\to$Eval$\to$Reflect$\to$Curate cycles) and 5$\times$ (20~calls, 5~cycles) on GPT-OSS-120B.
Table~\ref{tab:selfplay} reports the results.

\begin{table}[ht]
\centering
\small
\caption{Self-play budget ablation for ACE on GPT-OSS-120B, averaged across 3 schedules. More optimization \emph{degrades} quality.}
\label{tab:selfplay}
\setlength{\tabcolsep}{3pt}
\begin{tabular}{@{}l ccc c@{}}
\toprule
Budget & $\overline{\text{sat}}$ & Forg.$\downarrow$ & Refusal\% & Tok. \\
\midrule
1$\times$ (4 calls)  & 0.454 & 0.330 & 12.4\% & 1.62M \\
3$\times$ (12 calls) & 0.421 & 0.326 & 28.6\% & 2.39M \\
5$\times$ (20 calls) & 0.340 & 0.390 & 47.8\% & 2.82M \\
\midrule
Base LLM (0 calls)   & 0.630 & 0.179 & 0.0\% & 0.80M \\
\bottomrule
\end{tabular}
\end{table}

\paragraph{Key finding: more optimization accelerates failure.}
Mean satisfaction \emph{decreases} monotonically with budget ($0.454 \to 0.421 \to 0.340$), widening the gap to Base~LLM ($0.630$).
The mechanism is the Refusal Cascade (\S\ref{app:failures}): each cycle adds rules to the playbook, so 5$\times$ grows the playbook to 15.6K characters (vs.\ 9.1K at 1$\times$), exceeding the model's tolerance.
Refusal rates rise from 12\% to 48\%, directly explaining the satisfaction drop.

\paragraph{Schedule-dependent nuance:}
On the Rule-Only-15 schedule (6 rule-based types, deterministic evaluation), more cycles \emph{marginally help}: mean satisfaction rises $0.619 \to 0.624 \to 0.636$ and refusal rates actually \emph{decrease} ($10.9\% \to 9.1\% \to 6.8\%$).
Rule-based constraints produce unambiguous self-play signal---binary pass/fail on concrete specifications (word counts, keywords)---so the Reflector generates actionable rules that the Curator can integrate without bloat.
On LLM-judged schedules (Interleaved-20, Clustered-20), the same mechanism fails: vague constraint specifications (``adopt a persuasive narrative style'') produce noisy self-play verdicts, leading to verbose, hedge-laden playbook entries that accumulate into perceived contradictions.

\paragraph{Implication:}
The failure of proactive adaptation on complex schedules is \emph{structural}, not computational.
The bottleneck is not insufficient self-play optimization but distribution mismatch between the fixed synthetic self-play task and the diverse real evaluation prompts.
Scaling compute within this paradigm only accelerates artifact growth without improving transfer.

\section{Full-Constraints Information Ablation}
\label{app:full_constraints}

We test whether ACE's failure is due to an \emph{information gap}: in the standard protocol, \texttt{adapt()} sees only the current step's operation and a list of active type names.
In this ablation, we additionally provide the full text of \emph{all} currently active constraints, so the self-play pipeline evaluates against the same specifications the model will face at test time.
We run ACE on GPT-OSS-120B (the worst-affected backbone; standard ACE: $0.454$, base: $0.630$) across 3 schedules.

\begin{table}[ht]
\centering
\small
\caption{Full-constraints ablation for ACE on GPT-OSS-120B, averaged across 3 schedules. Closing the information gap recovers most of the deficit but yields zero net improvement over Base~LLM.}
\label{tab:full_constraints}
\setlength{\tabcolsep}{3pt}
\begin{tabular}{@{}l cccc c@{}}
\toprule
Condition & $\overline{\text{sat}}$ & Forg.$\downarrow$ & BWT$\uparrow$ & Switch$\uparrow$ & Tok. \\
\midrule
Base LLM       & \textbf{0.630} & \textbf{0.179} & \textbf{0.033} & 0.309 & 0.80M \\
ACE (full-ctx) & \underline{0.614} & \underline{0.203} & \underline{$-$0.032} & \textbf{0.321} & 2.26M \\
ACE (standard) & 0.454 & 0.330 & $-$0.180 & 0.289 & 1.62M \\
\bottomrule
\end{tabular}
\end{table}

\paragraph{Key finding: closing the information gap is necessary but not sufficient.}
Full-constraints ACE recovers 91\% of the standard ACE deficit ($0.454 \to 0.614$; gap to base narrows from $-0.176$ to $-0.016$) and nearly eliminates excess forgetting ($0.330 \to 0.203$, vs.\ base $0.179$).
The aggregate trajectory metrics show that the recovery is not confined to the initialization score.
However, despite this recovery, full-constraints ACE remains below Base on mean satisfaction and BWT and has greater forgetting; its edit-switch score is $0.012$ higher. The variant therefore approaches the Base satisfaction level at 2.8$\times$ the token cost without establishing a significant gain.

\paragraph{What drives the remaining gap:}
Refusal rates remain comparable to standard ACE (${\sim}13\%$ vs.\ ${\sim}12\%$; base: 0\%), indicating that the playbook representation itself---regardless of information quality---still causes occasional perceived contradictions.
The Gen$\to$Eval$\to$Reflect$\to$Curate pipeline introduces indirection: constraint knowledge must survive four LLM transformations before reaching the playbook, introducing noise and verbosity at each stage.

\paragraph{Implication:}
The information gap explains ACE's \emph{catastrophic} failure on GPT-OSS-120B, but the fundamental limitation is architectural: for constraint satisfaction tasks where the specification is self-contained, no amount of meta-cognitive machinery (reflection, curation, playbook management) improves upon directly presenting the constraints in the user prompt.
The constraints \emph{are} the optimal prompt.

\section{Gemma-4-26B-A4B Results}
\label{app:gemma}

As an additional model-family check, Table~\ref{tab:gemma426b} reports the complete trajectory-level and efficiency results for Gemma-4-26B-A4B.

\begin{table*}[ht]
\centering
\scriptsize
\caption{Trajectory-level and efficiency results for \textbf{Gemma-4-26B-A4B} (averaged over three schedules). Bold = best; underline = second best. These results reproduce the central trend on a third model family: Dynamic Cheatsheet differs from Base LLM by only $+0.001$, while every other adaptation method has lower mean satisfaction.}
\label{tab:gemma426b}
\setlength{\tabcolsep}{2.5pt}
\resizebox{\textwidth}{!}{%
\begin{tabular}{@{}l ccc ccc cc rr@{}}
\toprule
& \multicolumn{3}{c}{Satisfaction} & \multicolumn{3}{c}{Regression} & \multicolumn{2}{c}{Edit / Delete} & \multicolumn{2}{c}{Efficiency} \\
\cmidrule(lr){2-4} \cmidrule(lr){5-7} \cmidrule(lr){8-9} \cmidrule(lr){10-11}
Method & $\overline{\text{sat}}$ & Init. & Final & Forg.$\downarrow$ & Coll.$\downarrow$ & BWT$\uparrow$ & Switch$\uparrow$ & $\text{UF}_{\text{sus}}\!\uparrow$ & Lat.(s) & Tokens \\
\midrule
Base LLM         & \underline{0.677{\scriptsize$\pm$0.006}} & \underline{0.703{\scriptsize$\pm$0.131}} & \underline{0.639{\scriptsize$\pm$0.047}} & 0.163{\scriptsize$\pm$0.025} & 0.044{\scriptsize$\pm$0.034} & -0.075{\scriptsize$\pm$0.085} & \underline{0.325{\scriptsize$\pm$0.183}} & 0.702{\scriptsize$\pm$0.079} & \underline{45.7} & \textbf{394,911} \\
ICL              & 0.627{\scriptsize$\pm$0.037} & 0.675{\scriptsize$\pm$0.046} & 0.535{\scriptsize$\pm$0.179} & 0.167{\scriptsize$\pm$0.043} & 0.049{\scriptsize$\pm$0.024} & -0.056{\scriptsize$\pm$0.172} & 0.321{\scriptsize$\pm$0.203} & 0.701{\scriptsize$\pm$0.061} & \textbf{41.9} & \underline{654,828} \\
Dyn.\ Cheatsheet & \textbf{0.678{\scriptsize$\pm$0.017}} & 0.680{\scriptsize$\pm$0.051} & \textbf{0.679{\scriptsize$\pm$0.011}} & \textbf{0.134{\scriptsize$\pm$0.029}} & \textbf{0.031{\scriptsize$\pm$0.011}} & \textbf{0.092{\scriptsize$\pm$0.235}} & \textbf{0.334{\scriptsize$\pm$0.186}} & 0.717{\scriptsize$\pm$0.069} & 49.1 & 954,997 \\
ACE              & 0.504{\scriptsize$\pm$0.170} & 0.617{\scriptsize$\pm$0.040} & 0.470{\scriptsize$\pm$0.228} & 0.272{\scriptsize$\pm$0.083} & 0.056{\scriptsize$\pm$0.020} & -0.020{\scriptsize$\pm$0.227} & 0.288{\scriptsize$\pm$0.185} & 0.719{\scriptsize$\pm$0.062} & 53.1 & 854,704 \\
GEPA             & 0.658{\scriptsize$\pm$0.032} & 0.655{\scriptsize$\pm$0.018} & 0.617{\scriptsize$\pm$0.074} & \underline{0.161{\scriptsize$\pm$0.016}} & \underline{0.036{\scriptsize$\pm$0.017}} & \underline{0.027{\scriptsize$\pm$0.056}} & 0.311{\scriptsize$\pm$0.187} & \underline{0.749{\scriptsize$\pm$0.048}} & 57.9 & 683,111 \\
MIPROv2          & 0.660{\scriptsize$\pm$0.035} & \textbf{0.768{\scriptsize$\pm$0.090}} & 0.618{\scriptsize$\pm$0.109} & 0.172{\scriptsize$\pm$0.007} & 0.054{\scriptsize$\pm$0.033} & -0.026{\scriptsize$\pm$0.092} & 0.312{\scriptsize$\pm$0.202} & \textbf{0.767{\scriptsize$\pm$0.024}} & 51.5 & 663,811 \\
\bottomrule
\end{tabular}
}
\end{table*}

\end{document}